\documentclass[10pt,twocolumn,letterpaper]{article}

\usepackage{graphicx}
\usepackage{amsmath}
\usepackage{amssymb}
\usepackage{booktabs}

\usepackage{amsthm}
\theoremstyle{definition}

\usepackage{physics}
\usepackage{placeins}
\usepackage{float}

\usepackage{enumitem}
\setitemize{noitemsep,topsep=0.5pt,parsep=0.5pt,partopsep=0.5pt}
\setenumerate{noitemsep,topsep=0.5pt,parsep=0.5pt,partopsep=0.5pt}

\usepackage[breaklinks,colorlinks]{hyperref}

\usepackage[capitalize]{cleveref}
\crefname{section}{Sec.}{Secs.}
\Crefname{section}{Section}{Sections}
\Crefname{table}{Table}{Tables}
\crefname{table}{Tab.}{Tabs.}

\usepackage[
backend=biber,
sorting=ynt
]{biblatex}

\begin{document}

\title{Deep Signatures - Learning Invariants of Planar Curves}

\author{Roy Velich\\
Technion - Israel Institute of Technology\\
{\tt\small royve@campus.technion.ac.il}
\and
Ron Kimmel\\
Technion - Israel Institute of Technology\\
{\tt\small ron@cs.technion.ac.il}
}
\date{}
\maketitle

%%%%%%%%% ABSTRACT
\begin{abstract}
    We propose a learning paradigm for numerical approximation of differential invariants of planar curves.   
    Deep neural-networks' (DNNs) universal approximation properties are utilized to estimate geometric measures. 
    The proposed framework is shown to be a preferable alternative to axiomatic constructions.
    Specifically, we show that DNNs can learn to overcome instabilities and sampling artifacts and produce numerically-stable signatures for curves subject to a given group of transformations in the plane.
    We compare the proposed schemes to alternative state-of-the-art axiomatic constructions of group invariant arc-lengths and  curvatures. 
\end{abstract}

%%%%%%%%% BODY TEXT
%------------------------------------------------------------------------
\section{Introduction}
    \label{sec:intro}

    \paragraph{Differential Invariants and Signature Curves.}
        According to an important theorem by É. Cartan \cite{cartan1935}, two curves are related by a group transformation $g \in G$, that is, congruent curves with respect to $G$, if and only if their signature curves, with respect to the transformation group $G$, are identical. 
        This observation allows one to develop analytical tools to measure the equivalence of two planar curves, which has practical applications in various fields and tasks, such as computer vision, shape analysis, geometry processing, object detection, and more. 
        Planar curves are often extracted from images as boundaries of objects, or level-sets of gray-scale images. 
        The boundary of an object encodes vital information which can be further exploited by various computer vision and image analyses tools. 
        A common computer vision task requires one to find common characteristics between two objects in two different image frames. 
        This task can be naturally approached by representing image objects as signature curves of their boundaries. Since signature curves are proven to provide full solution for the equivalence problem of planar curves \cite[p.~183, Theorem 8.53]{olver_1999}, they can be further analyzed for finding correspondence between image objects.
        Signature curves are parametrized by differential invariants. 
        Therefore, in order to generate the signature curve of a given planar curve, one has to evaluate the required differential invariants at each point. 
        In practice, planar curves are digitally represented as a discrete set of points, which implies that the computation of differential invariant quantities, such as the curvature at a point, can only be numerically approximated using finite differences techniques.
        Since many important and interesting differential invariants are expressed as functions of high-order derivatives, their approximations are prone to numerical instabilities due to high sensitivity to sampling noise.
        For example, the equi-affine curvature, which is a differential invariant of the special-affine transformation group $\mathrm{SA}\left(2\right)$, is a fourth-order differential invariant, and its direct approximation using an axiomatic approach is practically infeasible for discrete curves, especially around inflection points. 
        % We propose to explore an alternative approach for producing numerically-stable signature curves for discretely sampled planar curves, by utilizing the power of neural-networks as universal approximators. 
        
        In this work, we examine the capabilities of machines, and specifically deep learning models, to implicitly compute invariant properties of various geometries. 
        We believe that by employing a learning approach to the classical concepts of transformation groups and differential invariants, one can derive alternative numerically stable methods for generating signatures curves of shapes under various symmetry groups.

    \paragraph{Contribution.}
        Inspired by the numerical approximation efforts of Calabi et al. \cite{Calabi98differentialand}, and the learning approaches introduced in \cite{pai2017learning, Lichtenstein2019}, our objective is to present a learning approach for generating signatures of planar curves with respect to the Euclidean, equi-affine, full-affine and projective transformation groups, by harnessing the power of deep neural-networks as robust universal approximators. 
        We believe that the proposed framework will lead to the following major contributions to the research of differential invariants and their various applications.
        \begin{itemize}
            \item A unified approach and environment for approximating differential invariants and generating signature curves with respect to various transformation groups, including the affine and projective transformation groups, for which, very few attempts to approximate their differential invariants in the discrete setting have been a made.
            \item A numerically stable approach for approximating differential invariants of discrete planar curves. 
            We show that neural networks are able to overcome the numerical instability which persists in the axiomatic approach, due to their empirical robustness to noise.
            So, for example, in the equi-affine case, our proposed learning based method is empirically applicable to all planar curves, and is not restricted to convex curves. 
            This relaxation is achieved by constraining the condition number of group transformations, as well as limiting the down-sampling rates of input curves used for training. 
            Without such restrictions, sampling, quantization, and reparametrization of the curves, would 
            practically eliminate the numerical information required for any representation model.
%            of a planar curve after  and therefore, diverge the network's optimization process from converging into a relevant invariant representation model. 
        \end{itemize}
        Moreover, we believe that the proposed framework will lay the foundations for future work on novel learning methods for differential invariants of higher dimensional geometric structures, such as surfaces embedded in $\mathbb{R}^3$.
        It would be beneficial to current research and development efforts in the field of computer vision and shape analysis, as well as to establish new research disciplines in the emerging field of geometric deep-learning.
%------------------------------------------------------------------------
\section{Related Work}
    \label{sec:related_work}
    \paragraph{Axiomatic Approximation of Differential Invariants.}
        The axiomatic approach for approximating the differential invariants of discrete planar curves, as presented by Calabi et al. in \cite{CALABI1996154,Calabi98differentialand}, is based on the joint invariants between points of a discrete planar curve. 
        As suggested by Calabi et al., given a transformation group $G$ and a planar curve $\mathcal{C}$, one can approximate the fundamental differential invariant of $G$ at a point $\vb{x} \in \mathcal{C}$, which is denoted by $\kappa\left(\vb{x}\right)$ and is known as the $G$-\textit{invariant curvature} at $\vb{x}$, by first interpolating a set $\mathcal{P}_{\vb{x}} \subset \mathcal{C}$ of sampled points from a small neighborhood around $\vb{x}$ with an auxiliary curve $\mathcal{C_{\vb{x}}}$, on which the differential invariant is constant by definition. Then, the constant differential invariant $\widetilde{\kappa}\left(\vb{x}\right)$ of $\mathcal{C_{\vb{x}}}$ is evaluated using the joint invariants of the points in $\mathcal{P}_{\vb{x}}$. Finally, $\widetilde{\kappa}\left(\vb{x}\right)$ is used as an approximation to $\kappa\left(\vb{x}\right)$.
        
        Since the approximation is calculated by joint invariants of $G$, its evaluation is unaffected by the action of a group transformation $g \in G$, and therefore, as the size of the mesh of points $\mathcal{P}_{\vb{x}}$ tends to zero, the approximation of the differential invariant at $\vb{x}$ converges to the continuous value. 
        For example, in the Euclidean case, the fundamental joint invariant of the groups $\mathrm{E}\left(2\right)$ and $\mathrm{SE}\left(2\right)$ is the Euclidean distance between a pair of points. By applying the approach suggested by Calabi et al., the Euclidean curvature $\kappa\left(\vb{x}\right)$ can be approximated by first interpolating a circle through $\vb{x_{i-1}}$, $\vb{x_i}$, and $\vb{x_{i+1}}$ (where $\vb{x_{i-1}}$ immediately precedes $\vb{x_{i}}$, and $\vb{x_{i+1}}$ immediately succeeds $\vb{x_{i}})$, and then exploit the Euclidean distances between those three points to calculate the constant curvature $\widetilde{\kappa}\left(\vb{x}\right)$ of the interpolated circle, using Heron's formula.
        
        The second differential invariant at a curve point $\vb{x}$, is defined as the derivative of the $G$-invariant curvature $\kappa$ with respect to the $G$-\textit{invariant arc-length element} $s$, and is denoted by $\kappa_s\left(\vb{x}\right)$. According to Calabi et al., $\kappa_s\left(\vb{x}\right)$ is approximated using finite differences, by calculating the ratio of the difference between $\widetilde{\kappa}\left(\vb{x}_{i-1}\right)$ and $\widetilde{\kappa}\left(\vb{x}_{i+1}\right)$ with respect to the distance between $\vb{x_{i-1}}$ and $\vb{x_{i+1}}$. 
        
        In their paper, Calabi et al. further explain how to approximate  differential invariants of planar curves with respect to the equi-affine group $\mathrm{SA}\left(2\right)$.
        In this case, the fundamental joint invariant is the triangle area defined by a triplet of points, and the auxiliary curve is a conic section, which possess a constant equi-affine curvature, and has to be interpolated through five curve points in the neighborhood of $\vb{x}$. 
        The main limitation of their method, in the equi-affine case, is that it is practically applicable only to convex curves, since the equi-affine curvature is not defined at inflection points.
    \paragraph{Learning-Based Approximation.}
        A more recent approach was introduced in \cite{pai2017learning}. 
        There, Pai et al. took a learning approach and used a Siamese convolutional neural network to learn the euclidean differential invariant from a dataset of discrete planar curves, which were extracted from the MPEG-7 database \cite{LateckiLakamperEckhardt2000}. 
        However, the scope of their work was lacking a module to approximate the Euclidean arc-length at each curve point, and therefore, a signature curve could not be generated. Moreover, they have not extended their work to learning differential invariants of less restrictive transformation groups, such as the affine and equi-affine groups. 
        In \cite{Lichtenstein2019}, Lichtenstein et al. introduced a deep learning approach to numerically approximate the
        solution to the Eikonal equation.
        They proposed to replace axiomatic local numerical solvers with a trained neural network that provides highly accurate estimates of local distances for a variety of different geometries and sampling conditions.
        In \cite{pmlr-v80-long18a}, Long et al. proposed a feed-forward deep network, called PDE-Net, to accurately predict dynamics of complex systems and to uncover the underlying hidden PDE models. Their main idea is to learn differential operators by learning convolution kernels, and apply neural networks, as well as other machine learning methods, to approximate the unknown nonlinear responses.
        
        In this paper, we present a unified learning approach to approximate differential invariants of planar curves with respect to any given transformation group.
    % \paragraph{Applications of Signatures Curves.}
    %      In \cite{HoffOlver2014},  Hoof and Olver use signature curves to automatically solve jigsaw puzzles. 
    %      They do that by representing the curved sides of the puzzle pieces as signature curves, and use this representation to determine whether two given pieces fit together. 
    %      In \cite{GrimShakiban2017}, Grim and Shakiban utilize the signature curves of skin lesion contours, extracted from medical images, to characterize melanomas and moles by showing that benign and malignant lesions have different global and local symmetry patterns in their signature curves.
    %  {\color{red} These two examples are somewhat esoteric and do not reflect the power of the method. 
    %  Consider removing them.
    %  }
%------------------------------------------------------------------------
\section{Mathematical Framework}
    \paragraph{Transformation Groups.}
        A \textit{transformation group} acting on a space $S$ is a set $\mathcal{F}$ of invertible maps $\varphi: S \rightarrow S$ such that $\mathcal{F}$ is a group with respect to function composition.
        
        The \textit{general linear group}, denoted by $\mathrm{GL}\left(n,\mathbb{R}\right)$, is defined as the set of all invertible linear transformations $T: \mathbb{R}^n \xrightarrow[]{} \mathbb{R}^n$. 
        An element $g \in \mathrm{GL}\left(n,\mathbb{R}\right)$ is called a \textit{group transformation}, or a \textit{group action}.
        
        The \textit{orthogonal group}, denoted by $\mathrm{O}\left(n\right) \subset \mathrm{GL}\left(n,\mathbb{R}\right)$, is defined as the set of all orthogonal  linear transformations (rotations, reflections and rotoreflections), and is explicitly given by $\mathrm{O}\left(n\right) = \left\{A \in \mathrm{GL}\left(n,\mathbb{R}\right) \vert A^TA = I\right\}$.
        
        The \textit{special orthogonal group}, denoted by $\mathrm{SO}\left(n\right) \subset \mathrm{O}\left(n\right)$, is defined as the set of all orientation-preserving orthogonal linear transformations (rotations), and is explicitly given by $\mathrm{SO}\left(n\right) = \left\{A \in \mathrm{O}\left(n\right) \vert \det\left(A\right) = 1\right\}$.
        
        The \textit{affine group}, denoted by $\mathrm{A}\left(n,\mathbb{R}\right) = \left\{\left(A,a\right) \vert A \in \mathrm{GL}\left(n,\mathbb{R}\right), a \in \mathbb{R}^n\right\}$, is a combination of applying first a linear transformation followed by a translation of the space by a fixed vector, and is explicitly given by the map $\forall x \in \mathbb{R}^n: x \mapsto Ax + a$.
        
        The \textit{special affine group}, which is also known as the \textit{equi-affine group}, denoted by $\mathrm{SA}\left(n\right) \subset \mathrm{A}\left(n\right)$, is defined as the set of all volume preserving affine transformations, and is explicitly given by $\mathrm{SA}\left(n\right) = \left\{\left(A,a\right) \in \mathrm{A}\left(n\right) \vert \det\left(A\right) = 1\right\}$.
        
        The \textit{Euclidean group}, denoted by $\mathrm{E}\left(n\right) \subset \mathrm{SA}\left(n\right)$, is defined as the set of all Euclidean-norm preserving affine transformations, and is explicitly given by $\mathrm{E}\left(n\right) = \left\{\left(A,a\right) \in \mathrm{A}\left(n\right) \vert A \in \mathrm{O}\left(n\right)\right\}$.
        
        The \textit{special Euclidean group}, denoted by $\mathrm{SE}\left(n\right) \subset \mathrm{E}\left(n\right)$, is defined as the set of all Euclidean-norm and orientation preserving affine transformations, and is explicitly given by $\mathrm{SE}\left(n\right) = \left\{\left(A,a\right) \in \mathrm{A}\left(n\right) \vert A \in \mathrm{SO}\left(n\right)\right\}$.
        
    % \paragraph{Invariants and Joint Invariants.}
    \paragraph{Invariants and Joint Invariants.}
        Let $G$ be a transformation group acting on a space $S$. An \textit{invariant} is a real-valued function $I: S \rightarrow \mathbb{R}$ that is unaffected by $G$, i.e., it satisfies $I\left(g \cdot x\right) = I\left(x\right)$ for all $g \in G$ and $x \in S$. It is also common to refer to $I$ as $G$-invariant.
        A \textit{joint invariant} is a $G$-invariant function $J: S_1 \times S_2 \times \hdots \times S_m \rightarrow \mathbb{R}$. In other words, $J$ is an invariant to the simultaneous action of $G$ on multiple copies of $S$, and specifically, it satisfies that $J\left(g \cdot x_1, g \cdot x_2, \hdots, g \cdot x_m\right) = J\left(x_1, x_2, \hdots, x_m\right)$ for all $g \in G$ and $x_i \in S_i$. 
        
        For example, the Euclidean norm $d\left(\vb{p}_1, \vb{p}_2\right) = \norm{\vb{p}_1 - \vb{p}_2}_2$, is a joint invariant function with respect to the Euclidean group $\mathrm{E}\left(2\right)$ and the special Euclidean group $\mathrm{SE}\left(2\right)$, since Euclidean transformations do not stretch space. Similarly, the triangle area $A\left(\vb{p}_1, \vb{p}_2,\vb{p}_3\right) = \frac{\abs{\left(\vb{p}_2 - \vb{p}_1\right) \times \left(\vb{p}_3 - \vb{p}_1\right)}}{2}$ is a joint invariant with respect to the special affine group $\mathrm{SA}\left(2\right)$, since equi-affine transformations preserve area.
        
    \paragraph{Prolongation of Group Actions.}
        The prolongation of group actions is a fundamental notion on which the definition of differential invariants is based on. When analyzing the action of a transformation group on a function, it is crucial to consider the induced action of the group on the function's derivatives. 
        This induced extension of the group action is also called the \textit{prolonged action} of the group, or \textit{prolongation}.
        
        Suppose a planar curve $\mathcal{C}$ is identified with the graph of function $f: \mathbb{R} \rightarrow \mathbb{R}$. Given a point $\left(x, f\left(x\right)\right) \in \mathcal{C}$, one can build an extended local coordinate-system $\vb{z}^{\left[n\right]} = \left(x, y^{\left[n\right]}\right) = \left(x, f\left(x\right), f^{\left(1\right)}\left(x\right), \hdots, f^{\left(n\right)}\left(x\right)\right)$, where $f^{\left(i\right)}\left(x\right)$ is the $i^{\mathrm{th}}$ derivative of $f$ at $x$, and $y^{\left[n\right]} = \left(f\left(x\right), f^{\left(1\right)}\left(x\right), \hdots, f^{\left(n\right)}\left(x\right)\right)$. In this setting, $y^{\left[n\right]}$ is called the \textit{$n^{\mathrm{th}}$ order prolongation of $f$ at $x$}.
        
        % and its transformed version $g \cdot \mathcal{C}$ are 
        % Let $g \in G$ be a group transformation acting on $\mathbb{R}^2$, and 
        %  and $\hat{f}: \mathbb{R} \rightarrow \mathbb{R}$, respectively.
        Similarly, the \textit{$n^{\mathrm{th}}$ order prolongation of group transformation} $g \in G$ is a map $\vb{g}^{\left[n\right]}: \mathbb{R}^{n+2} \rightarrow \mathbb{R}^{n+2}$, such that $\hat{\vb{z}}^{\left[n\right]} = \vb{g}^{\left[n\right]} \cdot \vb{z}^{\left[n\right]}$, where $\vb{z}^{\left[n\right]} = \left(x, y^{\left[n\right]}\right)$ and $\hat{\vb{z}}^{\left[n\right]} = \left(\hat{x}, \hat{y}^{\left[n\right]}\right)$. From a geometric point of view, the first prolongation of a group transformation $g$ can be identified with its action on the tangent lines to $\mathcal{C}$, the second prolongation is associated with its action on the osculating circles to $\mathcal{C}$, and so on.

    \paragraph{Differential Invariants.}
        Let $G$ be a transformation group acting on a $\mathbb{R}^2$. The $n^{\mathrm{th}}$ order \textit{differential invariant} of $G$ is an invariant function with respect to the $n^{\mathrm{th}}$ order prolongation of $G$. For example, suppose a planar curve $\mathcal{C}$ is identified with the graph of a function $f: \mathbb{R} \rightarrow \mathbb{R}$. The Euclidean curvature at the point $\left(x, f\left(x\right)\right) \in \mathcal{C}$ is given by,
        \begin{eqnarray}
            \label{eq:euclidean_curvature_invariant}
            \kappa &=& \frac{f''\left(x\right)}{\left(1 + f'\left(x\right)^2\right)^{3/2}}.
        \end{eqnarray}
        It is well known that $\kappa$ is an invariant function of the special Euclidean group $\mathrm{SE}\left(2\right)$, which makes it a second order differential invariant of $\mathrm{SE}\left(2\right)$. Moreover, the equi-affine curvature is a fourth order differential invariant with respect to $\mathrm{SA}\left(2\right)$, as discussed in \cite{olver_1995}.

    \paragraph{Invariant Differential Operators.}
        % A \textit{one-form} in $\mathbb{R}$ is given by the expression $\omega\left(x\right) = h\left(x\right)dx$, where $h: \mathbb{R} \rightarrow \mathbb{R}$ is an analytic function, and $dx$ is an infinitesimal change in $x$.
        
        % A fundamental example for a one-form, is given by the differential of an analytic function $f: \mathbb{R} \rightarrow \mathbb{R}$. The linear contribution to the change in $f$ due to an infinitesimal change $dx$ at a point $x$, is known as the \textit{differential} of $f$ at $x$, and is given by $df\left(x\right) = f'\left(x\right)dx$.
        
        Let $G$ be a transformation group acting on a space $S$. A one-form $\omega\left(x\right) = h\left(x\right)dx$ is called $G$-\textit{invariant} if and only if $h\left(x\right) = h\left(A\left(g,x\right)\right)\frac{\partial A}{\partial x}\left(g,x\right)$ for all $x \in S$ and $g \in G$, where $A\left(g,x\right) = g \cdot x$.
        
        For example, suppose a planar curve $\mathcal{C}$ is identified with the graph of a function $f: \mathbb{R} \rightarrow \mathbb{R}$. The Euclidean arc-length element at the point $\left(x, f\left(x\right)\right) \in \mathcal{C}$ is given by
            \begin{eqnarray}
                \label{eq:euclidean_arc_length_element}
                ds &=& \sqrt{1 + f'\left(x\right)^2}dx.
            \end{eqnarray}
        As shown in \cite{olver_1995}, $ds$ is an invariant one-form of the special Euclidean group $\mathrm{SE}\left(2\right)$.
        % Suppose a planar curve $\mathcal{C}$ is identified with the graph of a function $f: \mathbb{R} \rightarrow \mathbb{R}$, and let $G$ be a transformation group. 
        Moreover, suppose $ds\left(x\right) = p\left(x\right)dx$ is an invariant one-form, where $p$ is a real function of the prolongation of $f$ with respect to $G$. Then, an \textit{invariant differential operator} is defined as
        \begin{eqnarray}
            \label{eq:invariant_differential_operator}
            \frac{d}{ds} &=& \frac{1}{p} \  \frac{d}{dx}.
        \end{eqnarray}
        As discussed in \cite{olver_1995,olver_1999}, an invariant differential operator $\frac{d}{ds}$ maps differential invariants to differential invariants. 
        % The following example demonstrates it for $\mathrm{SE}\left(2, \mathbb{R}\right)$, 
        For example, by applying the special-Euclidean differential operator $\frac{d}{ds}$ to the special-Euclidean differential invariant $\kappa$, we get a new differential invariant, given by
        \begin{eqnarray}
            \label{eq:dk_ds_diff_inv}
            \kappa_s = \frac{d\kappa}{ds} = \frac{f'''\left(x\right) \left(1 + f'\left(x\right)^2\right) - 3 f''\left(x\right)^2 f'\left(x\right)}{\left(1 + f'\left(x\right)^2\right)^3}.
        \end{eqnarray}
        As can be seen by Equation (\ref{eq:dk_ds_diff_inv}), $\kappa_s$ is a third order differential invariant of $\mathrm{SE}\left(2\right)$.

        In general, as stated by \cite[p.~176, Theorem 8.47]{olver_1995}, an ordinary $r$-dimensional transformation group $G$ acting on $\mathbb{R}^2$ admits the following properties.
        \begin{enumerate}
            \item A unique differential invariant $\kappa$ (up to a function of it) of order $r - 1$, which is known as the $G$-\textit{invariant curvature}.
            \item A unique $G$-invariant one-from $ds = p\left(x\right)dx$ (up to a constant multiple) of order at most $r - 2$, which is known as the $G$-\textit{invariant arc-length element}.
            \item Any other differential invariant of $G$ is a function $I: \left(\kappa, \kappa_s, \kappa_{ss}, \hdots \right) \rightarrow \mathbb{R}$ of the $G$-invariant curvature $\kappa$ and its derivatives with respect to the $G$-invariant arc length $s$.
        \end{enumerate}
        For more information about differential invariants, invariant differential operators, and the fundamental theorem of differential invariants, see \cite{olver_1995,olver_1999,tuznik_olver_tannenbaum_2020,LieSophusTransGroup,LieSophusDiffInvariant,MovingCoframesI,MovingCoframesII,BrucksteinHoltNetravaliRichardson1992,BrucksteinKatzirLindenbaumPorat1992,BrucksteinNetravali1995,BrucksteinRivlinWeiss1996,CattLionsMorelColl1992,OlverSapiroTannenbaum1994,kimmel1996,KimmelZhangBronsteinABronsteinM2011}.
        %KimmelZhangBronsteinABronsteinM2009,

    \paragraph{Equivalence and Signature Curves.}
        Let $G$ be a transformation group acting on $\mathbb{R}^2$. Two planar curves $\mathcal{C}$ and $\hat{\mathcal{C}}$ are \textit{equivalent} with respect to $G$ if there exist $g \in G$ such that $\hat{\mathcal{C}} = g \cdot \mathcal{C}$. For example, two planar curves are equivalent with respect to $\mathrm{SE}\left(2\right)$ if one can be transformed into the other by a rotation and translation.
        
        A planar curve $\mathcal{C}$ is said to be $G$-\textit{ragular} if the differential invariants $\kappa$ and $\kappa_s$ are defined and smooth.
        The $G$\textit{-invariant signature curve} $S$ of a $G$-regular planar curve $\mathcal{C}$ is parametrized by the $G$-invariant curvature and its derivative with respect to the $G$-invariant arc-length, and is given by $S = \left\{\left(\kappa\left(\vb{x}\right), \kappa_s\left(\vb{x}\right)\right) \vert \vb{x} \in \mathcal{C}\right\}$.

        Signature curves can be used to solve the equivalence problem of planar curves for a general transformation group $G$, as stated by the \textit{equivalence of planar curves theorem} given in \cite[p.~183, Theorem 8.53]{olver_1995}, by which, two nonsingular curves $\mathcal{C}$ and $\hat{\mathcal{C}}$ are equivalent with respect to $G$ if and only if their signature curves $S$ and $\hat{S}$ are equal.

        Another approach to the solution of the equivalence problem of planar curves, as stated in \cite[Theorem 2-10]{guggenheimer2012differential}, is known as the \textit{traditional approach}, or as the \textit{traditional signature}. By that approach, given a transformation group $G$, a planar curve is characterized by its $G$-invariant curvature as a function of its $G$-invariant arc-length.
        Since the arc-length at a point can be measured with respect to an arbitrary fixed-point on the curve, the traditional signature exhibits a ``phase-shift" ambiguity.
        % This characterization leads to the following alternative equivalence theorem 
        % as presented in \cite[Theorem 2-10]{guggenheimer2012differential}, which states that,
        % \begin{theorem}
        %     Let $G$ be an ordinary transformation group acting on $\mathbb{R}^2$. Two non-singular curves $\mathcal{C}$ and $\hat{\mathcal{C}}$ are equivalent with respect to $G$ if and only if they have the same $G$-invariant curvature as a function of $G$-invariant arc-length, up to a phase-shift.
        % \end{theorem}
        % Note that the signature-curve approach has two main advantages over the traditional approach. First, the signature-curve approach is based on local information only, while the traditional approach requires to determine the arc-length (with respect to a fixed reference point) at each point by integration. Second, the signature-curve approach does not require a reference point for measuring arc-length, and thus does not possess a phase-shift ambiguity. However, from a numerical point of view, the signature-curve approach has a substantial disadvantage, since it requires to evaluate an additional derivative $\kappa_s$. 
        % For more elaborate discussion about equivalence of planar curves,  see \cite{olver_1995,olver_1999,guggenheimer2012differential}.
%------------------------------------------------------------------------
\section{Method}
    We introduce two simple neural-net architectures, along with appropriate training schemes and loss functions, for producing numerically stable approximations of $G$-invariant curvature and $G$-invariant arc-length, with respect to a group transformation $G$. 
    Those two architectures can be combined together to evaluate and plot the (traditional) $G$-invariant signature curve of a planar curve. 
    Given a planar curve $\mathcal{C}$ and a point $\vb{p} \in \mathcal{C}$ on the curve, the first architecture receives as an input a discrete sample of the local neighborhood of $\vb{p}$, and outputs the approximated $G$-invariant curvature at $\vb{p}$. Similarly, given two points $\vb{p}_1, \vb{p}_2 \in \mathcal{C}$ on the curve, the second architecture receives as an input a discrete sample of the curve section between $\vb{p}_1$ and $\vb{p}_2$, and outputs its approximated $G$-invariant arc-length.

    \paragraph{Dataset.}
        \label{sec:dataset}
        Our raw dataset consists of virtually unlimited number of discrete planar curves, which are extracted from random images scraped from the internet. In order to extract a discrete planar curve form an image, we first convert the image into a gray-scale image, then smooth it with a Gaussian blur filter, and finally, extract a closed level-curve that corresponds to an arbitrary intensity value.
        % , see Figure \ref{fig:level_curve_extraction}.
        % \begin{figure*}[htbp]
        %   \centering
        %   \includegraphics[width=\textwidth]{figures/level_curve_extraction@2x.png}
        %   \caption{Generation of a smooth planar curves by extraction of level-curves from raw images.}
        %   \label{fig:level_curve_extraction}
        % \end{figure*}

    \paragraph{Learning $G$-Invariant Curvature.}
        \label{sec:learning_curvature}
        The main observation that governs our training approach for learning the invariant curvature of a transformation group $G$, is that the neural-network should learn a representation that is not only unaffected by a group transformation $g \in G$, but which is also invariant to any given reparametrization of the input curve. 
        That way, the network will learn a truly invariant representation, which is not biased towards any specific discrete sampling scheme, and that encapsulate the local geometric properties of the curve.  
        Given a discrete planar curve $\mathcal{C}$, we denote by $\mathcal{C}\left(\mathcal{D}\right)$ its down-sampled version, sampled non-uniformly with $N_{\kappa}$ points according to a random non-uniform probability mass function $\mathcal{D}: \mathcal{C} \rightarrow \left[0,1\right]$. 
        Given two different non-uniform probability mass functions $\mathcal{D}_1$ and $\mathcal{D}_2$,
        one can refer to the down-sampled curves $\mathcal{C}\left(\mathcal{D}_1\right)$ and $\mathcal{C}\left(\mathcal{D}_2\right)$, as two different reparametrizations of $\mathcal{C}$ over the same interval $\mathcal{I}$.
        This motivates us to require that if a point $\vb{p} \in \mathcal{C}$ was drawn both in $\mathcal{C}\left(\mathcal{D}_1\right)$ and $\mathcal{C}\left(\mathcal{D}_2\right)$, then, a valid prediction model should output the same curvature value for $\vb{p}$, whether it was fed with either a neighborhood of $g_1 \cdot \vb{p}$ or a neighborhood $g_2 \cdot \vb{p}$ as an inference input, for any two group transformations $g_1, g_2 \in G$.
      
\begin{figure*}[htbp]
            \centering
            \includegraphics[width=\textwidth]{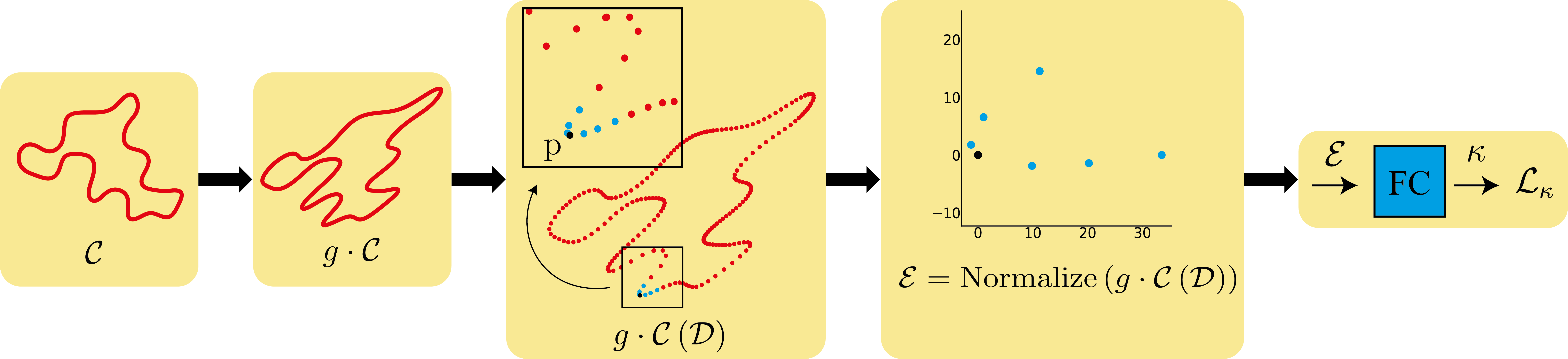}
            \caption{The $G$-Invariant curvature training scheme, demonstrated for an input example of a neighborhood of a point $\vb{x}$. In this illustration, $G$ is the equi-affine transformation group, and we have set $\mathcal{N}_{\kappa}=6$ and $N_{\kappa}$ equals to 30\% of the input curve points count (i.e., down-sampled by 70\%). 
            % (a) A random reference curve is transformed by a random group transformation $g \in G$. (b) The transformed curve is down-sampled with a random probability mass function $\mathcal{D}$, and a local neighborhood of a point $p$ is extracted as a training sample. (c) The extracted training sample is normalized. (d) The normalized input sample $\mathcal{E}$ is fed into the feed-forward neural network, which outputs the $G$-invariant curvature approximation $\kappa$ at $p$. The approximated curvature is passed into the loss function $\mathcal{L}_{\kappa}$. 
            %   In this illustration, $G$ is the equi-affine transformation group, and we have set $k=3$, and $N$ to be $10\%$ of the reference curve's points count.
            }
            \label{fig:curvature_tuplet}
        \end{figure*}

      %FFF  
    \paragraph{$G$-Invariant Curvature Training Scheme.}
        \label{sec:learning_curvature_training_scheme}
        Given a collection of discrete planar curves $\left\{\mathcal{C}_i\right\}_{i=1}^n$, which were generated by the method explained above, %in Section \ref{sec:dataset},
        our training scheme generates a training batch on the fly, where each batch is made of a set of training tuplets. 
        A curvature training tuplet $\mathcal{T}_{\kappa} = \left(\mathcal{E}_a, \mathcal{E}_p, \mathcal{E}_{n_1}, \hdots, \mathcal{E}_{n_{m_{\kappa}}}, \mathcal{E}_{n_{m_{\kappa}+1}}\right)$, is made up of collection of sampled neighborhoods, which are referred to as the \textit{anchor} ($\mathcal{E}_a$), \textit{positive} ($\mathcal{E}_p$) and \textit{negative} ($\left\{\mathcal{E}_{n_i}\right\}_{i=1}^{m_{\kappa}}$) examples. 
        The tuplet is generated by first drawing a random curve $\mathcal{C}$ and a random %working 
        point $\vb{p} \in \mathcal{C}$ on it. 
        Given two random group transformations $g_a, g_p \in G$ and two random non-uniform probability mass functions $\mathcal{D}_a$ and $\mathcal{D}_p$, both the anchor and positive examples, $\mathcal{E}_a$ and $\mathcal{E}_p$, are generated by selecting $2\mathcal{N}_{\kappa}$ adjacent points to $\vb{p}$ ($\mathcal{N}_{\kappa}$ consecutive points that immediately precedes $\vb{p}$, and additional $\mathcal{N}_{\kappa}$ consecutive points that immediately succeeds $\vb{p}$) from the transformed and down-sampled versions of $\mathcal{C}$ given by $g_a \cdot \mathcal{C}_a\left(\mathcal{D}_a\right)$ and $g_p \cdot \mathcal{C}_p\left(\mathcal{D}_p\right)$, respectively. 
        In other words, both $\mathcal{E}_a$ and $\mathcal{E}_p$ represent a local neighborhood of $\vb{p}$ sampled under different reparametrizations, which were transformed by group transformations taken from the same transformation group $G$. 
        The $i^{\mathrm{th}}$ negative example $\mathcal{E}_{n_i}$ is generated in a similar manner. 
        We draw an additional random point $\vb{p}_i \in \mathcal{C}$ from some local neighborhood of $\vb{p}$, such that $\vb{p}_i$ is in proximity to $\vb{p}$, but $\vb{p}_i \neq \vb{p}$. Then, we select again additional $2\mathcal{N}_{\kappa}$ adjacent points to $\vb{p}_i$ from a transformed and down-sampled version $g_{n_i} \cdot \mathcal{C}\left(\mathcal{D}_i\right)$ of $\mathcal{C}$, where $g_{n_i} \in G$ is a random group transformation, and $\mathcal{D}_i$ is a random probability mass function. 
        We train a multi-head Siamese feed-forward neural network by minimizing a \textit{tuplet loss function}, as described in \cite{9010708}, which is given by,
        \begin{eqnarray}
            \label{eq:loss_curvature}
            \mathcal{L}_{\kappa} = \log\left(1+\sum_{i=1}^{m+1} \exp\left(\left| \kappa_a - \kappa_p \right| - \left| \kappa_a - \kappa_{n_i} \right| \right) \right).
        \end{eqnarray}
        Where $\kappa_a$, $\kappa_p$ and $\kappa_{n_1}, \hdots, \kappa_{n_{{m_{\kappa}}+1}}$ are the neural-net's approximated curvature outputs for the input tuplet examples $\mathcal{E}_a$,  $\mathcal{E}_p$, and $\mathcal{E}_{n_1}, \hdots, \mathcal{E}_{n_{{m_{\kappa}}+1}}$, respectively. 
        Intuitively, in order to minimize (\ref{eq:loss_curvature}), the distance between the anchor $\kappa_a$ and the positive example $\kappa_p$ has to be minimized, and the distance between the anchor $\kappa_a$ and each negative example $\kappa_{n_i}$ has to be maximized.
        Note, that we eliminate any translation and rotational ambiguity by normalizing each input example $\mathcal{E}$. 
        The normalization is done by translating each example such that its midpoint is located at the origin, and then rotate it such that the ray that emanates from the sample's midpoint and passes through the sample's first point, is aligned with the positive x-axis.
        Moreover, note that we set the $\left(m_{\kappa}+1\right)^{\mathrm{th}}$ negative example to be the flipped version of the anchor example.
        By ``flipping'' we mean we convert a clock-wise orientation to a counter clock-wise one, and vice versa, 
        since we require the neural-network to distinguish between convex and concave curvatures. 
        See Figure \ref{fig:curvature_tuplet} for an elaborate visual demonstration of the $G$-invariant curvature training scheme.

    \paragraph{Learning $G$-Invariant Arc-Length.}
        Our approach for learning the $G$-invariant arc-length metric, follows the same motivation as described above,
        %in Section \ref{sec:learning_curvature}, 
        % You are in Section 4 :-)
        which requires the network to learn a property which is unaffected by any group transformation $g \in G$ and by any given reparametrization.
        The main difference, however, is that in this case, the network is aimed to learn an integrative property, rather than a differential and local one as in the case of the $G$-invariant curvature.
        Given a discrete planar curve $\mathcal{C}$ and two points $\vb{p}_1, \vb{p}_2 \in \mathcal{C}$ on it, we denote by $\mathcal{S}_1^2$ the curve section between $\vb{p}_1$ and $\vb{p}_2$. 
        Similarly to the above construction
        %presented in Section \ref{sec:learning_curvature}, 
        we denote by ${\mathcal{S}_1^2}\left(\mathcal{D}\right)$ the down-sampled version of $\mathcal{S}_1^2$, sampled non-uniformly with $N_s$ points according to a random non-uniform probability mass function $\mathcal{D}: \mathcal{S}_1^2 \rightarrow \left[0,1\right]$. We do force, however, that $\vb{p}_1, \vb{p}_2 \in {\mathcal{S}_1^2}\left(\mathcal{D}\right)$. 
        Given two different non-uniform probability mass functions $\mathcal{D}_1$ and $\mathcal{D}_2$, one can refer to two different down-sampled versions ${\mathcal{S}_1^2}\left(\mathcal{D}_1\right)$ and ${\mathcal{S}_1^2}\left(\mathcal{D}_2\right)$ of $\mathcal{S}_1^2$, as two different reparametrizations of $\mathcal{S}_1^2$ over the same interval $\mathcal{I}$.
        This construction motivates us to require that a valid prediction model should output the same arc-length value for $\mathcal{S}_1^2$, whether it was fed with either $g_1 \cdot {\mathcal{S}_1^2}\left(\mathcal{D}_1\right)$ or $g_2 \cdot {\mathcal{S}_1^2}\left(\mathcal{D}_2\right)$ as an inference input, for any two group transformations $g_1, g_2 \in G$.
    \paragraph{$G$-Invariant Arc-Length Training Scheme.}
        \label{sec:learning_arclength_training_scheme}
        Given a collection of discrete planar curves $\left\{\mathcal{C}_i\right\}_{i=1}^n$, which were generated by the above method,
        %explained in Section \ref{sec:dataset}, 
        our arc-length training scheme generates a training batch on the fly, where each batch is made of a set of training tuplets. 
        Let $\mathcal{C}$ be a random curve, and let $\vb{p}_1, \hdots, \vb{p}_{m_s}$ be $m_s$ random consecutive points on it, to which we refer as \textit{anchor points}. Given two group transformations $g_{\mathcal{A}}, g_{\mathcal{B}} \in G$, an arc-length training tuplet $\mathcal{T} = \left(\mathcal{A}, \mathcal{B}\right)$, is made up of two collections of sampled curve sections.
        The first collection is denoted by $\mathcal{A} = \left\{\mathcal{E}_i^{i+1}\right\}$ for all $i \in \left\{1,\hdots,m_s-1\right\}$, such that $\mathcal{E}_i^{i+1} = g_{\mathcal{A}} \cdot \mathcal{S}_i^{i+1}\left(\mathcal{D}_i^{i+1}\right)$ for a random non-uniform probability mass function $\mathcal{D}_i^{i+1}$.
        The second collection is denoted by $\mathcal{B} = \left\{\mathcal{E}_i^{j}\right\}$ for all $i,j \in \left\{1,\hdots,m_s\right\}$ such that $j - i > 1$, and where $\mathcal{E}_i^j = g_{\mathcal{B}} \cdot \mathcal{S}_i^j\left(\mathcal{D}_i^j\right)$ for a random non-uniform probability mass function $\mathcal{D}_i^j$.
        In other words, elements of $\mathcal{A}$ represent sampled curve sections which lay between two adjacent anchor points on a curve $\mathcal{C}$, which was transformed by a group transformation $g_{\mathcal{A}}$. On the other hand, members of $\mathcal{B}$ represent sampled curve sections which lay between two consecutive but non-adjacent anchor points on the same curve $\mathcal{C}$, which was transformed by a group transformation $g_{\mathcal{B}}$. Given any sampled curve section input $\mathcal{E}_i^j \in \mathcal{T}$, we denote by $s_i^j$ its associated neural-net's approximated arc-length output (note that $\mathcal{E}_i^j$ is also normalized as described above, before it is fed into the network). 
        We now use the notation presented above to define the arc-length loss function. By the definition of arc-length, for any $\mathcal{E}_i^j \in \mathcal{B}$, a valid arc-length prediction model would satisfy the following additive property requirement, given the sampled curve sections $\left\{\mathcal{E}_k^{k+1}\right\}_{k=i}^{j-1} \subset \mathcal{A}$,
        \begin{eqnarray}
            \label{eq:sum_req}
            \sum_{k=i}^{j-1} s_k^{k+1} &=& s_i^j.
        \end{eqnarray}
        Moreover, for any sampled curve section $\mathcal{E}_i^{i+1} \in \mathcal{A}$ such that $i+1 < m_s$, a valid arc-length prediction model would satisfy the following monotonic property requirement, for the sampled curve section $\mathcal{E}_i^{i+2} \in \mathcal{B}$,
        \begin{eqnarray}
            \label{eq:sum_req}
            s_i^{i+2} &>& s_i^{i+1}.
        \end{eqnarray}
        Following the motivation given above, we train a multi-head siamese feed-forward neural network by minimizing the following loss function, for a batch of arc-length training tuplets,
        {\small
        \begin{eqnarray}
            \label{eq:arclength_loss}
            \mathcal{L}_{s} =  \sum_{i = 1}^{m_s-2}
            \left ( \sum_{j=i+2}^{m_s} \left| s_i^j - \sum_{k=i}^{j-1} s_k^{k+1} \right|\,  +\, % \sum_{i=1}^{m-2} 
            \exp\left(s_i^{i+1} - s_i^{i+2}\right)
            \right ).
        \end{eqnarray}
        }
        % Note that in $(\ref{eq:arclength_loss})$, the first term can be thought of as the objective to be minimized, while the second term (the exponential) is a regularizing term, which prevents the network's weights to vanish to zero.
        \begin{figure*}[t]
          \centering
          \includegraphics[width=\textwidth]{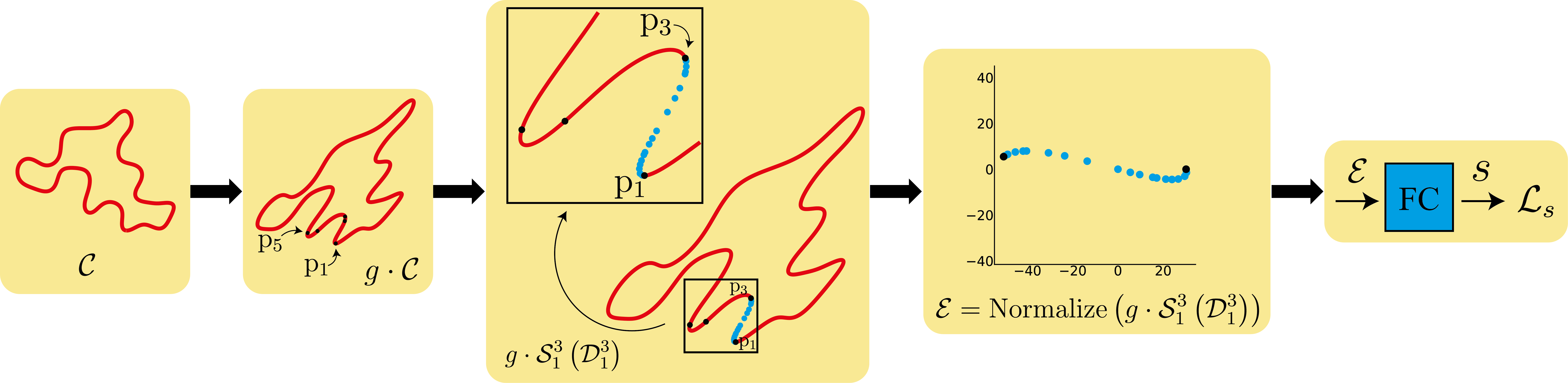}
           \caption{
           The $G$-Invariant arc-length training scheme, demonstrated for an input example of the curve section $\mathcal{S}_1^3$. In this illustration, $G$ is the equi-affine transformation group, and we have set $m_{s}=5$ and $N_{s}=40$. 
        %   (a) A random reference curve is transformed by a random group transformation $g \in G$, and $m$ random points are chosen. 
        %   (b) The curve section between $\vb{p}_1$ and $\vb{p}_3$ is down-sampled with a random probability mass function $\mathcal{D}_1^3$. 
        %   (c) The sampled curve section  training sample is normalized. 
        %   (d) The normalized input sample $\mathcal{E}$ is fed into the feed-forward neural network, which outputs the $G$-invariant arc-length approximation $s$ of $\mathcal{S}_1^3$. The approximated arc-length $s$ is passed into the loss function $\mathcal{L}_{s}$. 
        %   In this illustration, $G$ is the equi-affine transformation group, and we have set $m=5$ and $N=40$.
           }
           \label{fig:arclength_tuplet}
        \end{figure*}
%------------------------------------------------------------------------
\section{Experiments and Results}
    \paragraph{Datasets.}
        We have generated three independent collections of planar curves, each consists of tens of thousands of curves, which are used as the training, validation, and test datasets for both the $G$-invariant curvature and $G$-invariant arc-length neural-networks. 
        During training, we utilize a multi-core CPU environment to constantly generate training batches on the fly, as described above.
    % \paragraph{Hyper-Parameters}
    %     For the $G$-invariant curvature model, we have used $\mathcal{N}_{\kappa} = 3$, $N_{\kappa}$ set to 30\% of the input curve points count (i.e., down-sampling rate of 70\%), and $m_{\kappa} = 3$. We have used an LBFGS optimizer with learning-rate of 1, and batch size of 200000 training tuplets.
        
\begin{figure}[!h]
    \centering
    \includegraphics[width=1\linewidth]{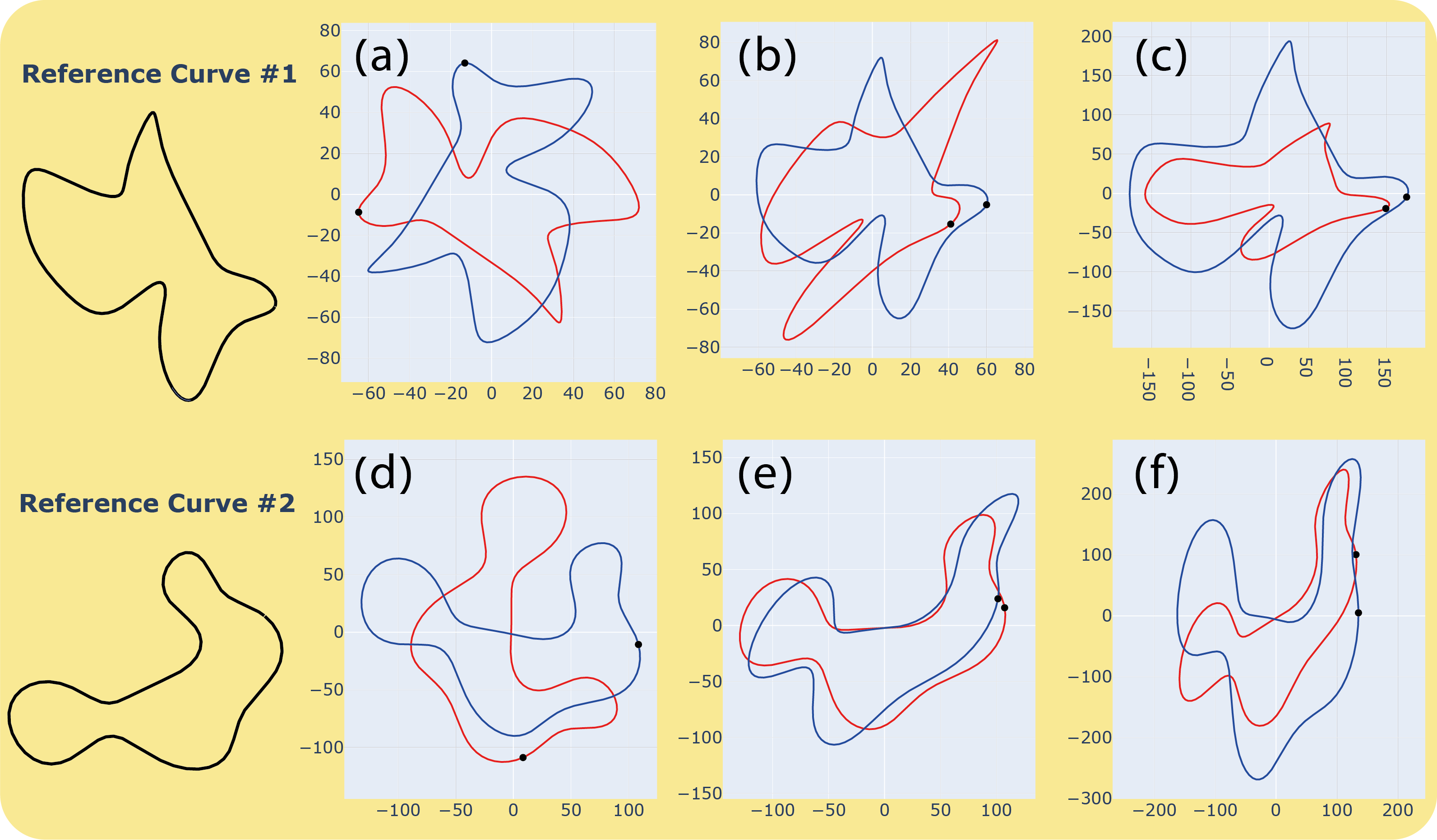}
    \caption{Two reference curves, each transformed by two random group transformations taken from $\mathrm{SE}\left(2\right)$ ((a) and (d)), $\mathrm{SA}\left(2\right)$ ((b) and (e)), and $\mathrm{A}\left(2\right)$ ((c) and (f)). The black dots indicate the arc-length measurement reference points (curve traversal is done in clock-wise order).}
    \label{fig:transformed_reference_curves}
\end{figure}

\begin{figure}[!h]
    \centering
    \includegraphics[width=1\linewidth]{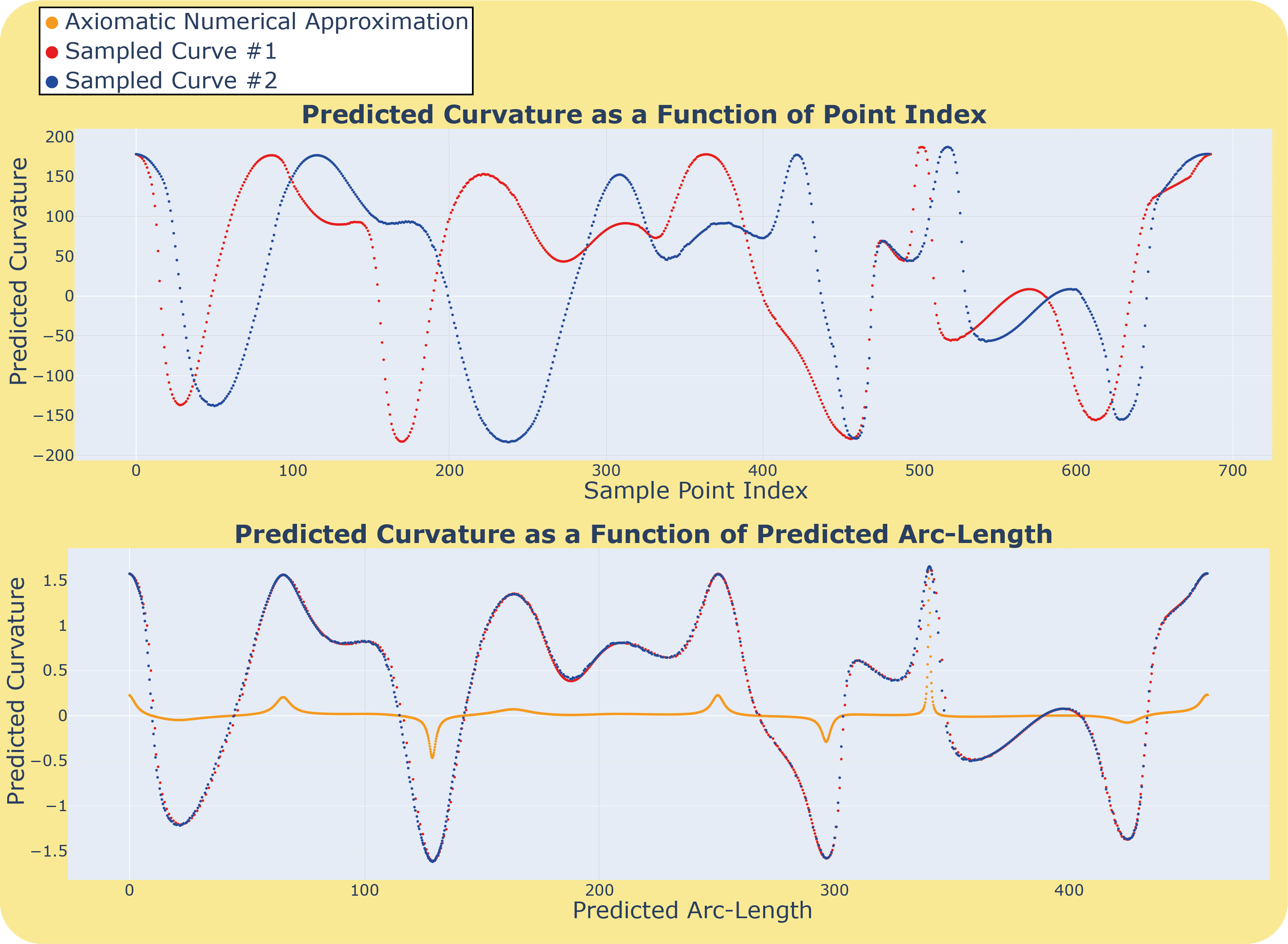}
    \caption{The estimated Euclidean signatures of the pair of curves denoted by \textbf{(a)} in Figure \ref{fig:transformed_reference_curves} (\textbf{down-sampled non-uniformly by 30\%}). Note, axiomatic signature is available only in the Euclidean case.}
    \label{fig:signature_a}
\end{figure}

\begin{figure}[!h]
    \centering
    \includegraphics[width=1\linewidth]{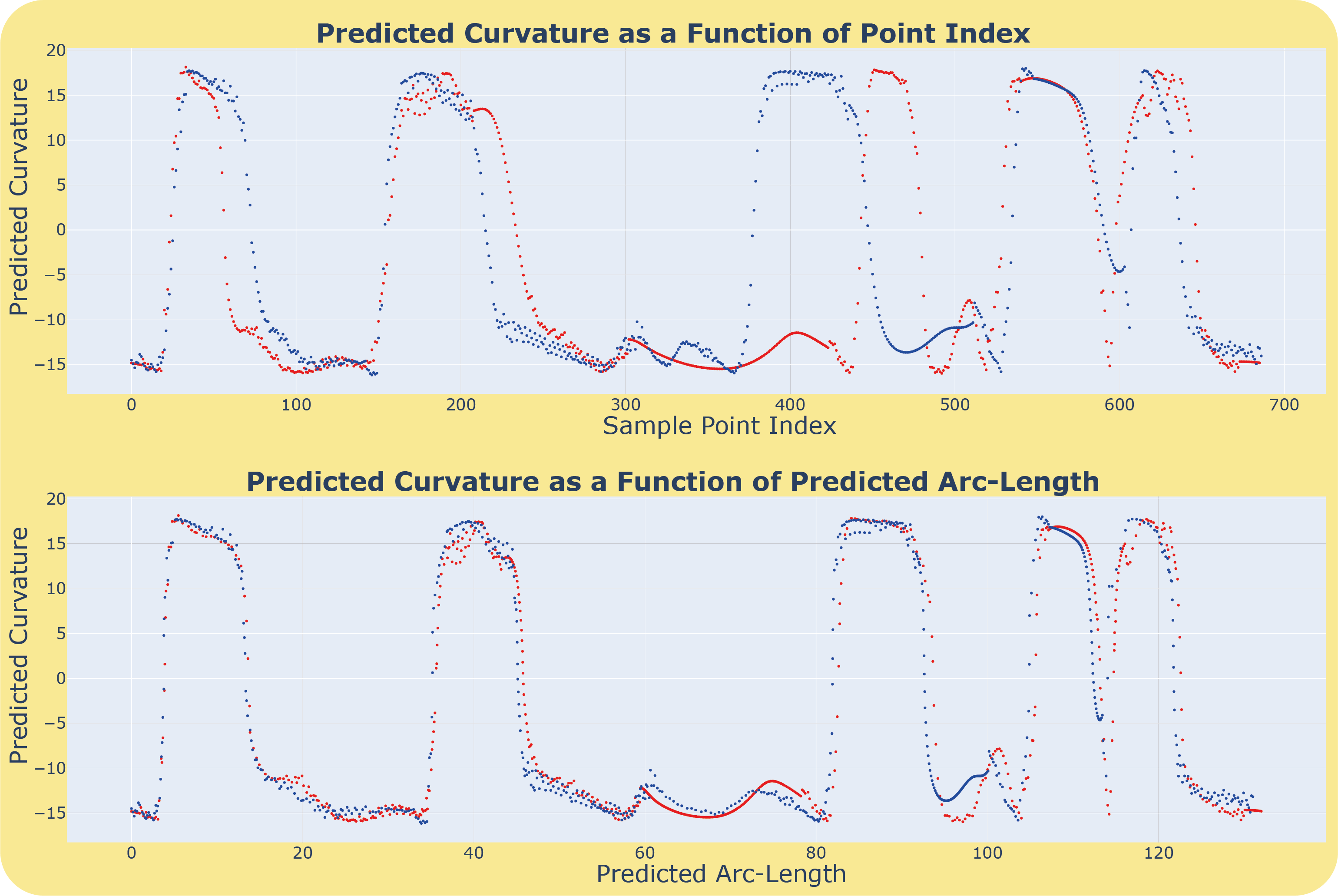}
    \caption{The estimated equi-affine signatures of the pair of curves denoted by \textbf{(b)} in Figure \ref{fig:transformed_reference_curves} (\textbf{down-sampled non-uniformly by 30\%}).}
    \label{fig:signature_b}
\end{figure}

    %     For the $G$-invariant arc-length model, we have used $\mathcal{N}_{s}$ and $m_{s}$ equal to 5. We have used an LBFGS optimizer with learning-rate of 0.1, and batch size of 150000 training tuplets.
    \paragraph{Measuring Arc-Length Between Adjacent Points.}
        Given a trained $G$-invariant arc-length neural-network $M: \mathbb{R}^{N_s \times 2} \rightarrow \mathbb{R}$, that is designed to evaluate the arc-length of curve-section samples comprised of $N_s$ points, we can exploit it to evaluate the arc-length element between two adjacent points $\vb{p}_i, \vb{p}_{i+1} \in \mathcal{C}$. First, we define by $\mathcal{S}_i = \left\{\vb{p}_{i-N_s+1},\hdots,\vb{p}_i\right\}$ the curve-section that is comprised of $N_s$ consecutive points, of which $\vb{p}_i$ is an end-point. Next, given an arbitrary $i-N_s+1 < j < i+1$, we define by $\mathcal{S}_{i+1} = \left\{\vb{p}_{i-N_s+1},\hdots, \vb{p}_{j-1}, \vb{p}_{j+1}, \hdots,\vb{p}_{i+1}\right\}$ the curve-section that is comprised of $N_s$ consecutive points that excludes $\vb{p}_j$, of which $\vb{p}_{i+1}$ is an end-point. Finally, The arc-length between $\vb{p}_i$ and $\vb{p}_{i+1}$ is given by $M\left(\mathcal{S}_{i+1}\right) - M\left(\mathcal{S}_{i}\right)$.
    \paragraph{Qualitative Evaluation.}
        We evaluate our $G$-invariant curvature and $G$-invariant arc-length approximation models qualitatively, by plotting the learned invariant parametrization (also known as the ``traditional" invariant signature, as defined above) of non-uniformly down-sampled planar curves. Given a planar curve $\mathcal{C}$ taken from the test dataset, we first generate two down-sampled and transformed versions of $\mathcal{C}$, given by $\widetilde{\mathcal{C}}_1 = g_1 \cdot \mathcal{C}\left(\mathcal{D}_1\right)$ and $\widetilde{\mathcal{C}}_2 = g_2 \cdot \mathcal{C}\left(\mathcal{D}_2\right)$, for two random probability mass functions $\mathcal{D}_1$ and $\mathcal{D}_2$, and two random group transformations $g_1, g_2 \in G$.
        In order to evaluate the $G$-invariant curvature for $\widetilde{\mathcal{C}}_i$, we iterate over each point $\vb{p} \in \widetilde{\mathcal{C}}_i$, extract a local neighborhood sample of $\vb{p}$ in $\widetilde{\mathcal{C}}_i$, normalize it, and feed it into the $G$-invariant curvature neural-network.
        Similarly, in order to compute the $G$-invariant arc-length for $\widetilde{\mathcal{C}}_i$, we first choose an arbitrary point $\vb{p}^r \in \mathcal{C}$ as a reference point. Then, we iterate again over each point $\vb{p} \in \widetilde{\mathcal{C}}_i$, and evaluate the $G$-invariant arc-length between every two adjacent points along the curve section between $\vb{p}$ and $\vb{p}^r_i = g_i \cdot \vb{p}^r$ (the section's interior points are taken only from $\widetilde{\mathcal{C}}_i$). The summation of all adjacent-points' arc-length evaluations is the $G$-invariant arc-length approximation at $\vb{p}$.
        Next, we plot a graph of the evaluated $G$-invariant curvature as a function of sample-point index, for both $\widetilde{\mathcal{C}}_1$ and $\widetilde{\mathcal{C}}_2$, and show that the two graphs are not aligned.
        Finally, we plot another graph of the evaluated $G$-invariant curvature as a function of the evaluated $G$-invariant arc-length, for both $\widetilde{\mathcal{C}}_1$ and $\widetilde{\mathcal{C}}_2$, and show that, this time, the two graphs are aligned as expected. In practice, we have trained separate curvature and arc-length approximation models for the Euclidean, equi-affine and affine transformation groups. Experimentation results are available in Figures \ref{fig:transformed_reference_curves}, \ref{fig:signature_a}, \ref{fig:signature_b}, \ref{fig:signature_c}, \ref{fig:signature_d}, \ref{fig:signature_e}, \ref{fig:signature_f}, and in the supplementary material.
    \paragraph{Observations.}
        Given a planar curve $\mathcal{C}$, both the Euclidean and equi-affine arc-length values can be analytically calculated, for any $\vb{x} \in \mathcal{C}$ \cite{Kimmel2004}. Interestingly, we empirically observe that in those two cases, our trained arc-length models learn a representation which depends linearly on the ground-truth corresponding analytic expressions. So, for example, given a curve-section $\mathcal{S}_i^j$, our Euclidean arc-length model estimation is empirically given by $M_{\text{euclidean}}\left(\mathcal{S}_i^j\right) = c \cdot \sum_{k=i}^{j-1} \norm{\vb{p}_k - \vb{p}_{k+1}}_2$, for some constant $c \in \mathbb{R}$. By eliminating the linear constant $c$, we recover the ground-truth arc-length estimation model, as demonstrated in Figures \ref{fig:signature_a}, \ref{fig:signature_d}.
\section{Limitations}
    We acknowledge the following limitations of the proposed approach.
    \begin{itemize}
        \item Badly conditioned group transformations, and severely under-sampled curve-sections and point neighborhoods, may introduce numerical instability to our models, especially with respect to affine transformations.
        \item Currently, we only generate ``traditional" curve signatures, that is, curvature as a function of arc-length, which exhibit phase-shift ambiguity.
        \item Our current database is based on planar curves which were extracted as level-sets of gray-scale images. 
        A richer dataset should include planar curves from various independent sources.
    \end{itemize}
    
%___________________________________
\section{Conclusion and Future Work}
    We proposed a self-supervised approach, using multi-head Siamese neural networks, for the approximation of differential invariants of planar curves, with respect to a given transformation group $G$. 
    We showed that one can combine our models to produce almost identical signatures for two equivalent curves, with respect to $G$. 
    We further empirically demonstrate the robustness of our method to reparametrization, by generating aligned signatures for non-uniformly down-sampled equivalent curves.
    
    For future work, we plan to design models for the approximation of higher order differential invariants directly, such as $\kappa_s$, which will pave the way to the generation of unambiguous signature-curves of planar curves, which are based on local invariants only. 
    Moreover, we aim to design learning models for approximation of higher dimensional differential invariants \cite{olver2009}, such as the torsion of space curves, or the principal curvatures of surfaces embedded in Euclidean space.

\begin{figure}[!h]
    \centering
    \includegraphics[width=1\linewidth]{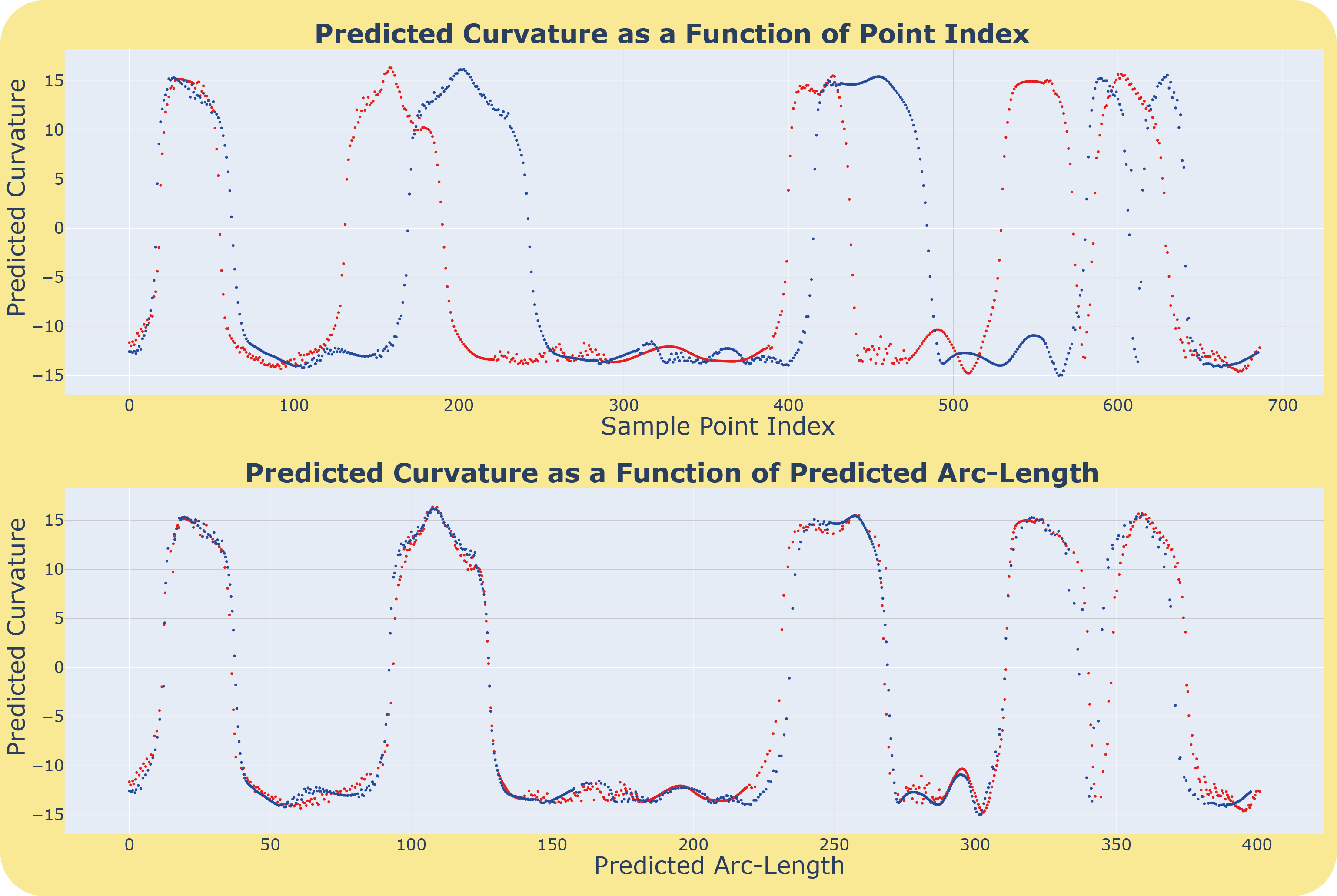}
    \caption{The estimated affine signatures of the pair of curves denoted by \textbf{(c)} in Figure \ref{fig:transformed_reference_curves} (\textbf{down-sampled non-uniformly by 30\%}).}
    \label{fig:signature_c}
\end{figure}

\begin{figure}[!h]
    \centering
    \includegraphics[width=1\linewidth]{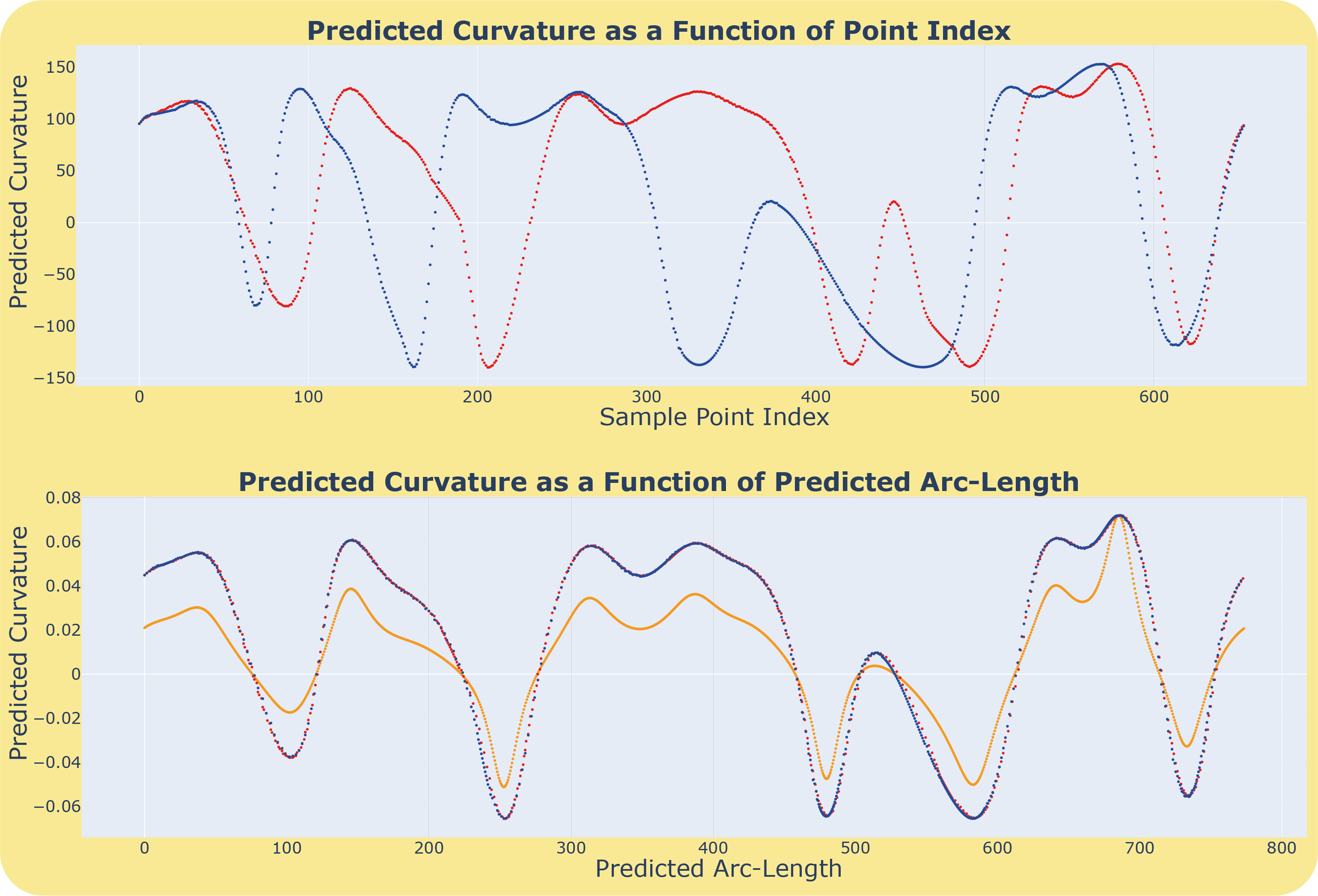}
    \caption{The estimated Euclidean signatures of the pair of curves denoted by \textbf{(d)} in Figure \ref{fig:transformed_reference_curves} (\textbf{down-sampled non-uniformly by 50\%}).}
    \label{fig:signature_d}
\end{figure}

\begin{figure}[!h]
    \centering
    \includegraphics[width=1\linewidth]{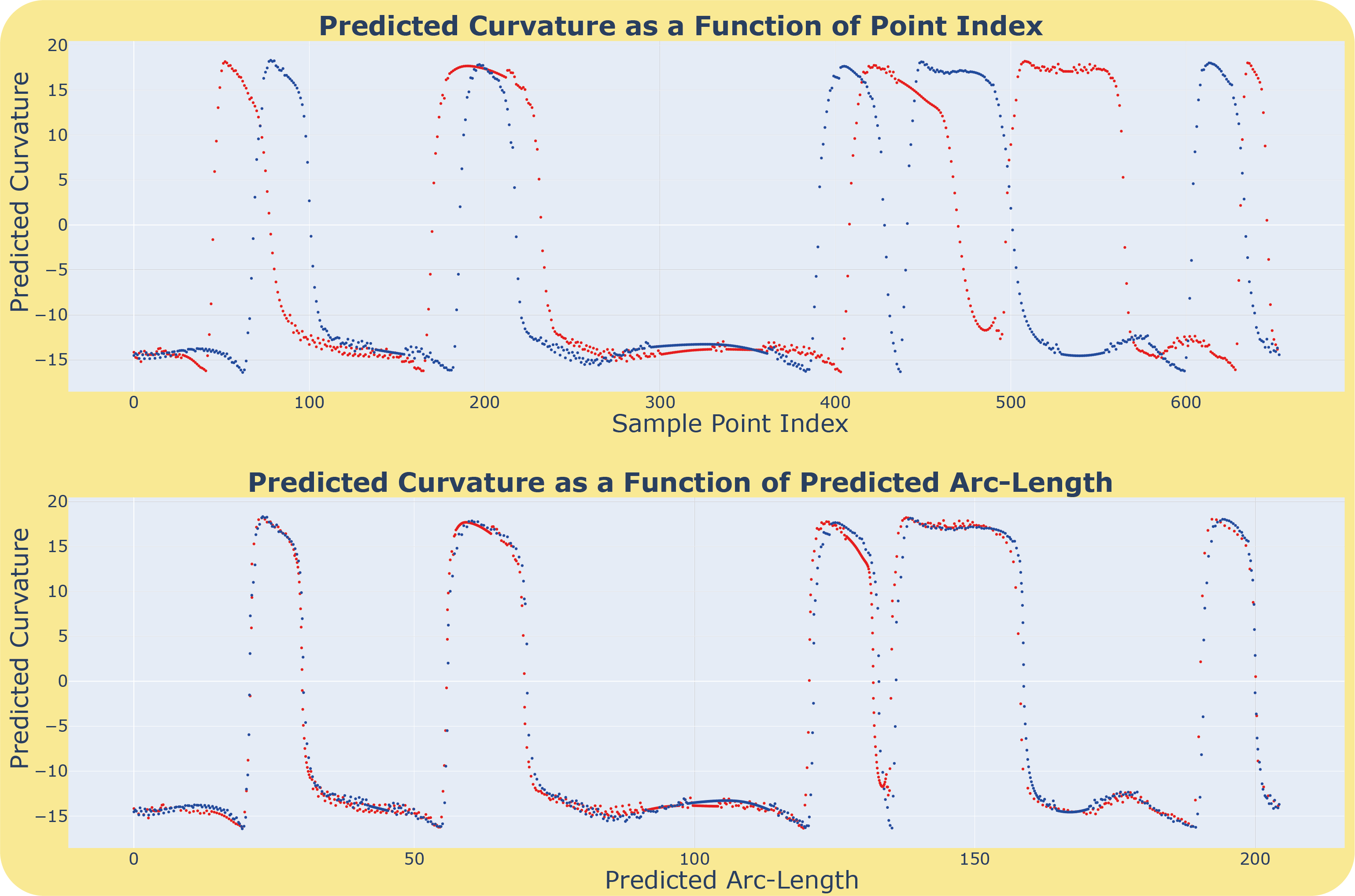}
    \caption{The estimated equi-affine signatures of the pair of curves denoted by \textbf{(e)} in Figure \ref{fig:transformed_reference_curves} (\textbf{down-sampled non-uniformly by 50\%}).}
    \label{fig:signature_e}
\end{figure}

\begin{figure}[p]
    \centering
    \includegraphics[width=1\linewidth]{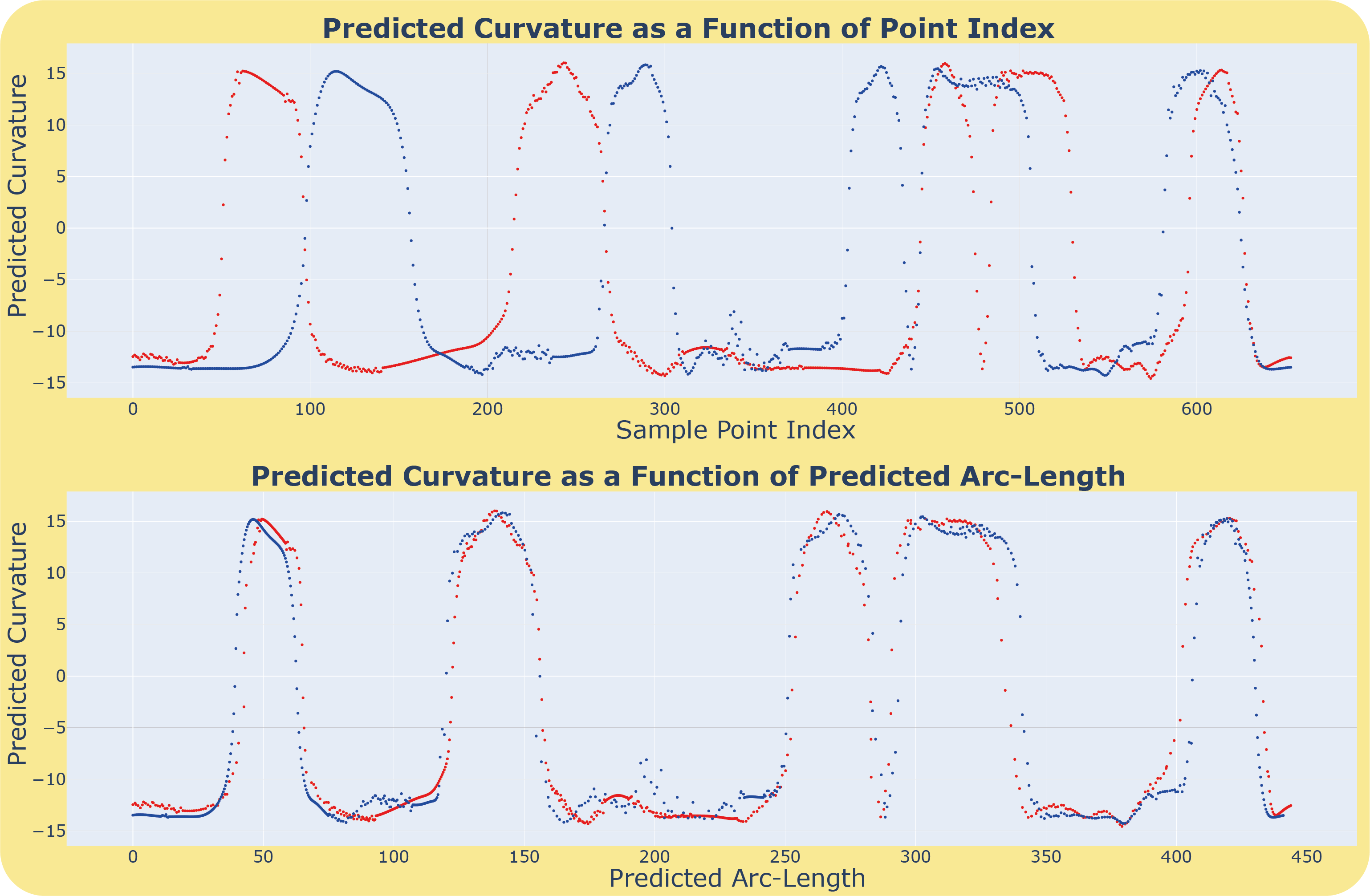}
    \caption{The estimated affine signatures of the pair of curves denoted by \textbf{(f)} in Figure \ref{fig:transformed_reference_curves} (\textbf{down-sampled non-uniformly by 50\%}).}
    \label{fig:signature_f}
    \vspace{128in}
\end{figure}

\FloatBarrier
%------------------------------------------------------------------------
%%%%%%%%% REFERENCES
\printbibliography

\end{document}